\RequirePackage{fix-cm}
\documentclass[twocolumn, natbib]{svjour3}          
\smartqed  
\usepackage{graphicx}
\usepackage{amsmath}
\usepackage{amssymb}
\usepackage{bbm}
\DeclareMathOperator{\sign}{sign}
\usepackage{booktabs}
\usepackage{url}
\usepackage[draft,colorlinks=true,pagebackref=true,breaklinks=true,bookmarks=false]{hyperref} 

\usepackage{paralist}
\usepackage[normalem]{ulem}

\newcommand{\abc}[1]{\textcolor{black}{#1}} 

\newcommand{\zq}[1]{\textcolor{black}{#1}} 
\newcommand{\zqb}[1]{\textcolor{black}{#1}} 

\newcommand{\zqq}[1]{\textcolor{black}{#1}} 

\newcommand{\ZQ}[1]{\textcolor{black}{}} 
\newcommand{\ZQNOTE}[1]{\textcolor{black}{}} 

\newcommand{\CUT}[1]{}

\journalname{International Journal of Computer Vision}

\begin{document}

\title{3D Crowd Counting via Geometric Attention-guided Multi-View Fusion}



\author{Qi Zhang         \and
        Antoni B. Chan 
}


\institute{Qi Zhang \at
              Shenzhen University, City University of Hong Kong \\
              \email{qzhang364-c@my.cityu.edu.hk}           
           \and
           Antoni B. Chan \at
              City University of Hong Kong \\
              \email{abchan@cityu.edu.hk}           
}

\date{Received: date / Accepted: date}

\maketitle

\begin{abstract}

\abc{Recently multi-view crowd counting using deep neural networks has been proposed to enable counting in large and wide scenes using multiple cameras.}
The current methods project the camera-view features to the average-height plane of the 3D world, and then fuse the projected multi-view features to predict a 2D scene-level density map on the ground \abc{(i.e., birds-eye view)}.
Unlike the previous research, we consider the variable height of the people in the 3D world and propose to solve the multi-view crowd counting task through 3D feature fusion with 3D scene-level density maps, instead of the 2D density map on the ground-plane.
  Compared to 2D fusion, the 3D fusion extracts more information of the people along the $z$-dimension (height), which helps to address the scale variations across multiple views. The 3D density maps still preserve the 2D density maps property that the sum is the count, while also providing 3D information about the crowd density.
  \zqb{Furthermore, instead of using the standard method of copying the features along the view ray in the 2D-to-3D projection, we propose an attention module based on a height estimation network, which forces each 2D pixels to be projected to one 3D voxel along the view ray.}
  We also explore the projection consistency among the 3D prediction and the ground-truth in the 2D views  
  to further enhance the counting performance. The proposed method is tested on the synthetic and real-world multi-view counting datasets and achieves better or comparable counting performance to the state-of-the-art.
\end{abstract}

\section{Introduction}
\begin{figure}[t]
\centering
   \includegraphics[width=0.8\columnwidth]{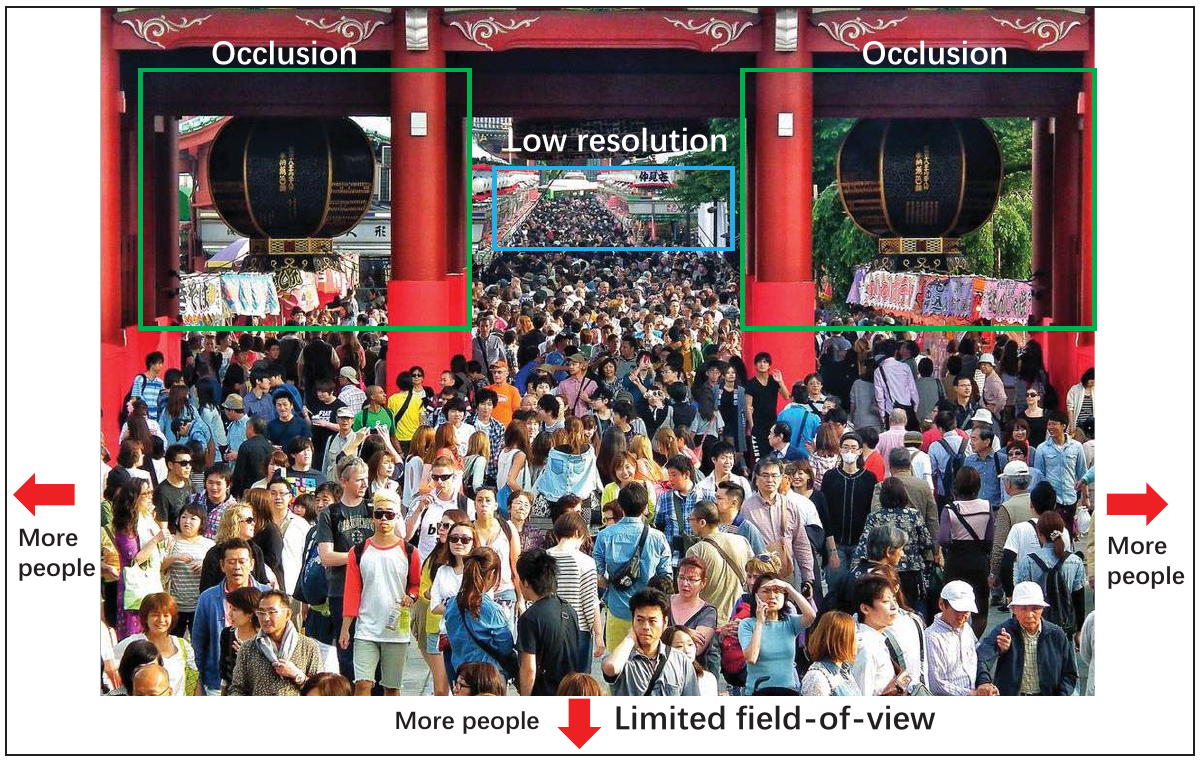}
   \caption{An example for the limitation of single-view counting for large and wide scenes: limited field-of-view, low resolution and severe occlusion. This image is from the ShanghaiTech dataset \citep{zhang2015cross}.
   }
\label{fig:example}
\end{figure}

Single-view crowd counting has been studied extensively and has achieved promising performance on the existing counting datasets \citep{zhang2015cross,Idrees2013Multi,zhang2016single,Chan2008Privacy,Chen2012Feature,idrees2018composition}. However, in real-world applications, there are several situations where single-view counting cannot perform well (e.g., see Fig.~\ref{fig:example}): 1) the scene is too wide (such as a park or a football match stadium) where a single camera's field-of-view is limited and cannot cover the whole scene; 2) the scene is too long (such as the underground train platform) where the people who are far away from the camera have very low resolution and the counting performance drops on these people; 3) the scene contains many obstacles, such as vehicles, building structures, \emph{etc.}, where many people are heavily or totally occluded. Under the 3 conditions, the current single-view counting methods are inaccurate, because many people are miscounted due to limited field-of-view, low resolution and severe occlusion.

\begin{figure*}[t]
\centering
   \includegraphics[width=0.8\textwidth]{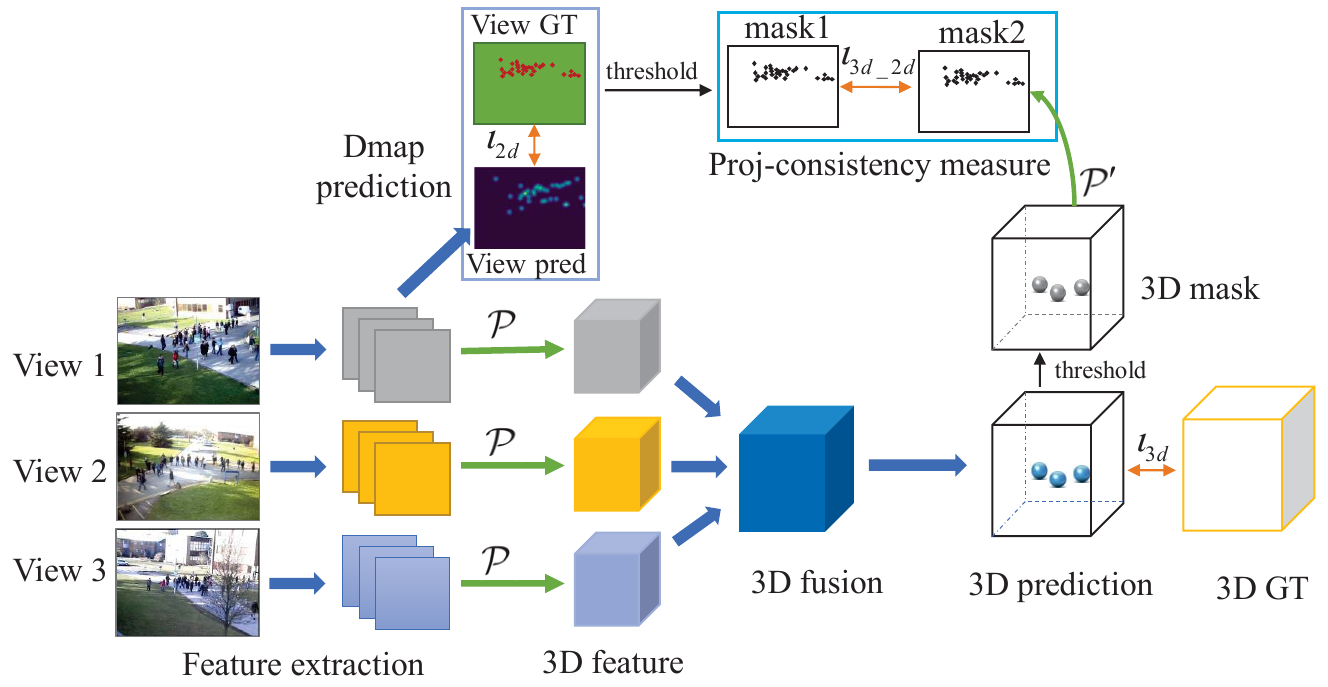}
   \caption{The pipeline of 3D crowd counting. Single-view features are extracted and then projected to the 3D world on multiple height planes. The projected 3D features are concatenated and fused to output the 3D density map prediction (loss $l_{3d}$). Each camera-view prediction branch decodes the 2D features to obtain the 2D camera-view predictions (loss $l_{2d}$). Finally, the 3D prediction is back-projected to each camera-view, and the the projection consistency between the camera-view ground-truth and the back-projected prediction 
   is measured (loss $l_{3d\_2d}$).}
\label{fig:pipeline}
\end{figure*}

To address the aforementioned situations, multiple cameras should be deployed, and the multi-view information should be fused to enhance the counting performance for complicated scenes. A few works \citep{li2012people,Ryan2014Scene,Tang2014Cross,Ge2010Crowd} have considered multi-view counting, but the hand-crafted features and the foreground extraction steps limit their performance. 
Recently, a CNN-based multi-view counting method \citep{zhang2019wide} was proposed, which improved the counting performance on wide-area scenes \abc{through the use of view fusion with deep neural networks.}
\cite{zhang2019wide} projected the multi-view information (camera-view density maps or feature maps) to a 2D plane (at the height of a person) in the 3D scene, and then fused them to predict the 2D scene-level density maps on the ground-plane. Several fusion methods are considered, including late and early fusion, and a multi-view multi-scale early fusion model (MVMS), which handles both the inter- and intra-view scale variations.

The disadvantage of the 2D-to-2D projection used by \cite{zhang2019wide} is that the features of the same person from different views may not line up correctly due to the approximation that all features come from the same height in the 3D world. Clearly this is not true for features extracted from the head and feet of a person.
To address this problem, in this paper, we propose to use 3D projection and 3D feature fusion to perform the multi-view counting task. Our proposed method consists of the following components (see Fig.~\ref{fig:pipeline}): 1) \emph{single view feature extraction and density map prediction}: 2D single-view features are extracted and then decoded to the 2D density map predictions; 2) \emph{2D-to-3D projection}: a differentiable projection layer together with the camera parameters are used to perform the fixed 3D projection from image plane to 3D world; 3) \emph{3D fusion and prediction}: the projected multi-view 3D features are fused using 3D CNN layers to predict the 3D scene-level density map \CUT{and $z$-dim full-size 3D filters are adopted to handle the scale variation issue}; 4) \emph{Projection consistency between the 3D prediction and 2D views}: the 3D prediction is back-projected to each camera view and a loss between the camera view prediction and back-projected 3D prediction is added to enhance the multi-view counting performance.

\zqb{In our approach, the 3D projection provides more information about each person along the $z$-dimension (height).
}
\zqb{We notice that the 2D-to-3D projection used in many DNNs-based multi-view fusion methods \cite{zhang2019wide, kar2017learning, Iskakov2019Learnable} form the 3D feature map by copying 2D features along the view-ray in the 3D volume.
This violates \abc{the principle}
that each 2D pixel should only come from one 3D voxel.
To consider this geometric constraint between 3D and 2D space, we propose a geometric attention mechanism along the projection view ray by estimating the probable pixel height in the 3D world.}
In addition, a 3D density map with 3D Gaussian kernels is used to represent the crowd in the scene, instead of a 2D scene-level density map. The 3D density map can provide more information about the crowd in 3D, e.g., the elevation of the crowd. 
\CUT{The prediction of the 3D density map is also used as an 
intermediate representation to enforce the projection consistency across the camera views, which can further improve the multi-view counting performance.
}

In summary, the main contributions of our paper are as follows:
\begin{compactitem}
  \item We propose an end-to-end DNNs-based 3D multi-view counting method that predicts 3D density maps, which provides information about the crowd in 3D.
  \item \zq{Unlike  previous methods, we use 3D projection and 3D fusion, which can help to deal with the scale variation issue without the scale selection module}.
  \item \zqb{We consider the geometry in the projection and propose an attention module along the view ray for better 2D-to-3D projection.}
  \item The projection consistency between the camera-view ground-truth and the back-projected 3D predictions is explored to further boost the multi-view counting performance. The proposed method can achieve better or comparable counting performance to the state-of-art.
\end{compactitem}

The remainder of this paper is organized as follows.
In Section 2, we review the existing single-view and multi-view counting methods, as well as 3D object reconstruction using deep neural networks (DNNs).
In Section 3, we introduce the proposed 3D multi-view counting method including the 3D fusion and geometric attention module.
In Section 4, we show the experiments settings and results, and conclude in Section 5.

\section{Related Work}
In this section, we review the existing single-view and multi-view counting methods. We also review 3D object reconstruction using deep neural networks (DNNs).
A preliminary conference version of this paper appears in AAAI \citep{zhang20203d}.
This journal paper extends the conference version in terms of following aspects:
1) a geometric-guided attention module along the view ray is proposed to reduce the feature redundancy caused by the previous 2D-to-3D projection steps;
2) updated 3D counting performance on the DukeMTMC dataset, which is better than in the previous conference version;
3) More experiments on more multi-view counting datasets to better validate the proposed method.

\subsection{Single-view counting}
Single-view crowd counting has achieved satisfactory performance on the existing counting datasets, especially the DNN-based density map estimation methods \citep{sindagi2020learning, jiang2020attention, bai2020adaptive, yang2020reverse}. \cite{zhang2015cross} proposed a single-column CNN framework to directly estimate the density maps from the images. Scale variations due to perspective changes 
 is a critical issue in the crowd counting task, which can limit the performance, and many 
 methods have been proposed to handle scale variations \citep{boominathan2016crowdnet,zhang2016single,sam2017switching,Kang2018Crowd,onoro2016towards}.
 \cite{sindagi2017generating} proposed the contextual pyramid CNN (CP-CNN) to incorporate global and local context information in the crowd counting framework. Furthermore, extra information and more sophisticated networks were utilized to further improve the counting performance \citep{idrees2018composition,Wang2019Learning,ranjan2018iterative,cao2018scale,li2018csrnet,liu2018decidenet,shen2018crowd,Jiang2019Crowd,Liu2019Context}.
 \cite{kang2017incorporating} proposed an adaptive convolution neural network (ACNN) by utilizing the context (camera height and angle) as side information in the counting framework.
 \cite{shi2019revisiting} integrated the perspective information to provide additional knowledge of the person scale change in an image.
 \cite{Lian2019Density} proposed a regression guided detection network (RDNet) for RGB-D crowd counting.
 \cite{Liu2019Recurrent} proposed Recurrent Attentive Zooming Network to zoom high density regions for higher-precision counting and localization.
Besides methods relying on density map for supervision, methods utilizing other forms of supervision are also proposed, such as classification \citep{Xiong_2019_ICCV}, local counting map \citep{liu2020adaptive}, and dot maps \citep{wang2020distribution, ma2019bayesian}.

The single-view based counting methods cannot handle the situations when the camera view cannot capture the whole scene,  the occlusions in the scene are too severe, or the people are in low resolution due to long distance from the camera. Therefore, multiple cameras should be adopted to improve the counting performance for these wide-area scenes.

\subsection{Multi-view counting}

Traditional multi-view counting methods can be divided into 3 categories: detection/tracking based \citep{dittrich2017people,li2012people,ma2012reliable,Maddalena2014people}, regression based \citep{Ryan2014Scene,Tang2014Cross}, and 3D cylinder based methods \citep{Ge2010Crowd}. These multi-view counting methods have several limitations. First, they need to utilize foreground extraction techniques to segment the crowd from background. Therefore, the effectiveness of the foreground extraction step limits the final counting performance. Second, hand-crafted features are used both in the people detection or crowd count regression. The hand-crafted features lack representation ability, which reduces the robustness and the performance of the methods. Third, these methods are mainly developed and tested on the PETS2009 \citep{ferryman2009pets2009}, which is a multi-view dataset with small crowd numbers and staged crowd behavior.

Recently, \cite{zhang2019wide} proposed a DNN-based multi-view counting method and a new larger multi-view counting dataset, CityStreet. This multi-view multi-scale (MVMS) DNN first extracts camera-view information (density maps or features), and then projects them to the average-height plane in the 3D scene with the given camera parameters. Next, the projected features are fused and decoded to predict the scene-level density maps (on the average height plane). Our proposed method differs from MVMS in the several aspects: 1) instead of using average-height projection \zqq{(2D-2D projection)}, we use multi-height projection \zqq{(2D-3D projection)}, which spatially aligns the person's features (e.g., head, body and feet features) in 3D, making it easier to find the geometric correspondence across views (See Fig. \ref{fig:3dprojection}); 2) we predict a 3D crowd density map, generated using 3D Gaussian kernels, which provides distribution of the crowd in 3D space; 3) the 3D density map prediction is back-projected to each camera view, and compared with the 2D ground-truth density map of the camera view, which defines a projection consistency loss for improving the accuracy.

\subsection{DNN-based 3D reconstruction}

Our 3D crowd counting method is related to a few works on DNN-based 3D object reconstruction and human pose estimation. \cite{yan2016perspective} proposed an unsupervised single-view 3D object reconstruction method utilizing the projection loss defined by the perspective transformation.
\cite{choy20163d} proposed a 3D RNNs architecture to gradually refine the multi-view 3D object reconstruction step-by-step.
\cite{kar2017learning} leveraged the 3D geometry constraint of features through projection and unprojection operations in the 3D reconstruction method.
\cite{huang2018deepmvs} presented DeepMVS to 
 produce a set of plane-sweep volumes and then predict high-quality disparity maps.
\cite{Iskakov2019Learnable} proposed two learnable triangulation methods for 3D human pose estimation: algebraic triangulation and volumetric aggregation.

\par
It can be observed that the existing DNNs-based 3D reconstruction methods are mainly focused on the reconstruction of a single object (e.g., see the ShapeNet \citep{chang2015shapenet} and IKEA \citep{lpt2013ikea} datasets, or human pose). Our proposed method can also be regarded as predicting a 3D representation from multiple viewpoints, but differs from the existing DNN-based 3D object reconstruction in several aspects. First, the proposed method can do more than \zq{3D shape reconstruction}, because the 3D Gaussian kernels can represent the crowd's 3D distribution as well as indicate the crowd count. Second, unlike the previous 3D object reconstruction, the proposed method targets at the scene-level representation (all people in the scene), not only a single object. Furthermore, we exploit the relationship between the 2D camera-view density maps and 3D scene-level density maps to obtain a projection consistency loss.

\begin{figure}[t]
\centering
   \includegraphics[width=\columnwidth]{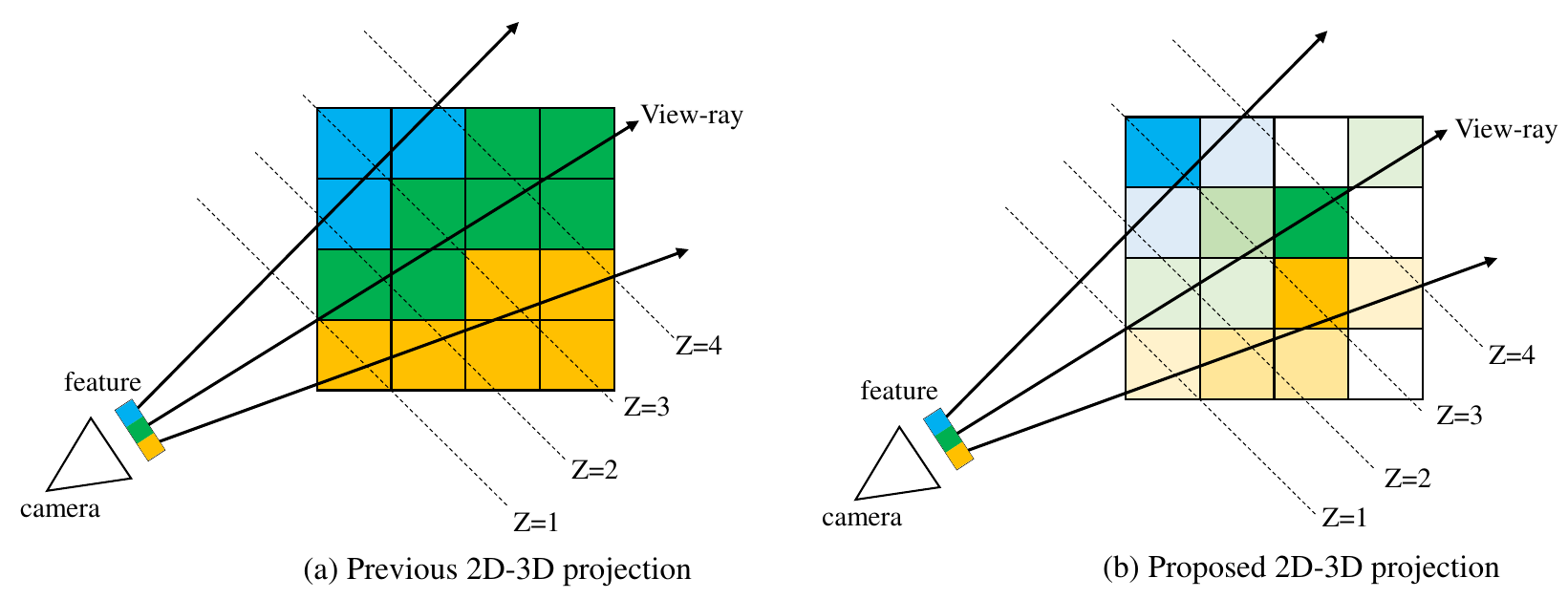}
   \caption{
   The 2D-3D projection process. (a) previous projection forms the 3D grids by copying the 2D features along the view-ray to the 3D grids with intersection; and (b) the proposed geometric attention guided 2D-3D projection process, which forces one 2D feature to be projected to one 3D voxel and other grids are suppressed. 
   On the right, lighter colors  indicates lower probability of the 3D voxel coming from the corresponding 2D feature along the view-ray.
 }
\label{fig:attention}
\end{figure}

\subsection{\zqb{2D-to-3D projection in CNNs}}
For DNNs-based multi-camera fusion tasks, such as 3D reconstruction from multi-view images \citep{kar2017learning} or 3D human pose estimation \citep{Iskakov2019Learnable},  the projection modules are used to project multi-view 2D features to the same 3D space for fusion (\zqq{see Fig. \ref{fig:attention}a}).
According to the ideal camera pin-hole calibration model,
\abc{a 2D image coordinate corresponds to  multiple 3d locations along the view ray, since the $Z$-dimension coordinate is unknown.}
To handle this ill-posed problem, several approximation methods are utilized to form the 2D-to-3D projection operation in DNNs.

\cite{kar2017learning} proposed the differentiable unprojection operation to project the 2D single-view image features into 3D space via copying the 2D features to the 3D voxels along the view ray. Many 3D reconstruction and 3D pose estimation methods have adopted this projection operation for multi-view fusion. However, the 3D voxels along the same view ray share the same features, which introduces redundancy in the 3D feature representation.
\cite{sitzmann2019deepvoxels} added an occlusion reasoning module for handling occlusion in the 3D-to-2D projection.
\cite{zhang2019wide} were the first to utilize DNNs for multi-view crowd counting. In their work, the multi-view features were projected to the average-height plane in the 3D space for fusion, which means each pixel's height ($Z$-dimension in their calibration model) is assumed to be average height of a person (1.75m). In the conference version of this paper \citep{zhang20203d}, the 2D features are projected to multiple heights to handle the wrong projection for other people body parts as in \citep{zhang2019wide}. However, the feature redundancy caused by the feature copying still exists, which results in fusion ambiguity across camera views.
To alleviate the problem of feature redundancy in the projection step,
we propose an attention module along the view ray to reduce the influence of repeated features  in the 3D feature map on the multi-view fusion and decoding nets.

\begin{table}[t]
\centering
\begin{tabular}{ll}
\scriptsize
\begin{tabular}{|c|c|}
\hline
\multicolumn{2}{|c|}{Single-view branch} \\ \hline
Layer         & Filter      \\ \hline
\multicolumn{2}{|c|}{Feature extraction} \\ \hline
conv 1             & $16\! \times\! 1\! \times\!  5\!  \times\!  5$     \\ 
conv 2             & $16\!  \times\!  16\!  \times\!  5\!  \times\!  5$    \\ 
pooling   & $2\!  \times\!  2\!  $         \\ 
conv 3             & $32\!  \times\!  16\!  \times\!  5\!  \times\!  5$   \\ 
conv 4             & $32\!  \times\!  32\!  \times\!  5\!  \times\!  5$ \\ 
pooling   & $2\!  \times\!  2\!  $          \\ \hline

\multicolumn{2}{|c|}{Density map prediction} \\ \hline

conv 5             & $64\!  \times\!  32\!  \times\!  5\!  \times\!  5$ \\ 
conv 6             & $32\!  \times\!  64\!  \times\!  5\!  \times\!  5$ \\ 
conv 7             & $1\!  \times\!  32\!  \times\!  5\!  \times\!  5$  \\ \hline
\end{tabular}
&
\scriptsize
\begin{tabular}{|c|c|}
\hline
\multicolumn{2}{|c|}{3D Fusion module}  \\ \hline
Layer & Filter     \\ \hline
   concatenation   &  - \\ 
3D conv 1     & $32\!  \times\!  n\!   \times\!  5\!  \times\!  5\!  \times\!7$   \\ 
3D conv 2     & $64\!  \times\!  32\!  \times\!  5\!  \times\!  5\!  \times\!7$  \\
3D conv 3     & $128\! \times\!  64\!  \times\!  5\!  \times\!  5\!  \times\!7$ \\
3D conv 4     & $64\!  \times\!  128\! \times\!  5\!  \times\!  5\!  \times\!7$ \\
3D conv 5     & $32\!  \times\!  64\!  \times\!  5\!  \times\!  5\!  \times\!7$ \\
3D conv 6     & $32\!  \times\!  32\!  \times\!  5\!  \times\!  5\!  \times\!7$ \\
3D conv 7     & $1\!   \times\!  32\!  \times\!  5\!  \times\!  5\!  \times\!7$ \\

\hline
\end{tabular}
\end{tabular}
\caption {The layer settings for the camera-view feature extraction and density map prediction branches (left), and the 3D fusion module (right). 
The filter dimensions are output channels, input channels, and filter size $w_0\!  \times\!  h_0\!  \times\!d_0$ ($d_0=1$ in 2D conv layers). }
\label{table:layer_setting}
\end{table}

\section{3D Counting via 3D Projection and Fusion}

We follow the setup of multi-view counting from \cite{zhang2019wide}: we assume fixed cameras with known intrinsic and extrinsic camera parameters, and synchronized camera frames across views. In contrast to \cite{zhang2019wide}, which is based on 2D ground-plane density maps, we generate 3D ground-truth density maps 
by convolving the 3D ground-truth annotations with fixed-width 3D Gaussian kernels. The 3D ground-truth annotation coordinates are calculated from the 2D view annotations and people correspondence across views (see Section 4 for more details).

In this section, we introduce the end-to-end DNN-based 3D crowd counting method 
consisting of the following stages: 1) \emph{Single view feature extraction and density map prediction}: 2D single-view features are extracted and  decoded to predict a 2D camera-view density map; 2) \emph{2D-to-3D projection}: a differentiable projection layer using the camera parameters projects the camera-view feature maps from the image plane to multiple height-planes in the 3D world; 3) \emph{3D fusion and prediction}: the projected multi-view 3D features are fused to predict the 3D density maps using 3D CNN layers; 4) \emph{Projection consistency between the 3D prediction and 2D views}: the 3D prediction is back-projected to each camera view, 
and then compared with the corresponding camera-view 2D ground-truth using loss to refine the 3D prediction.

\begin{figure}[t]
\centering
   \includegraphics[width=0.95\columnwidth]{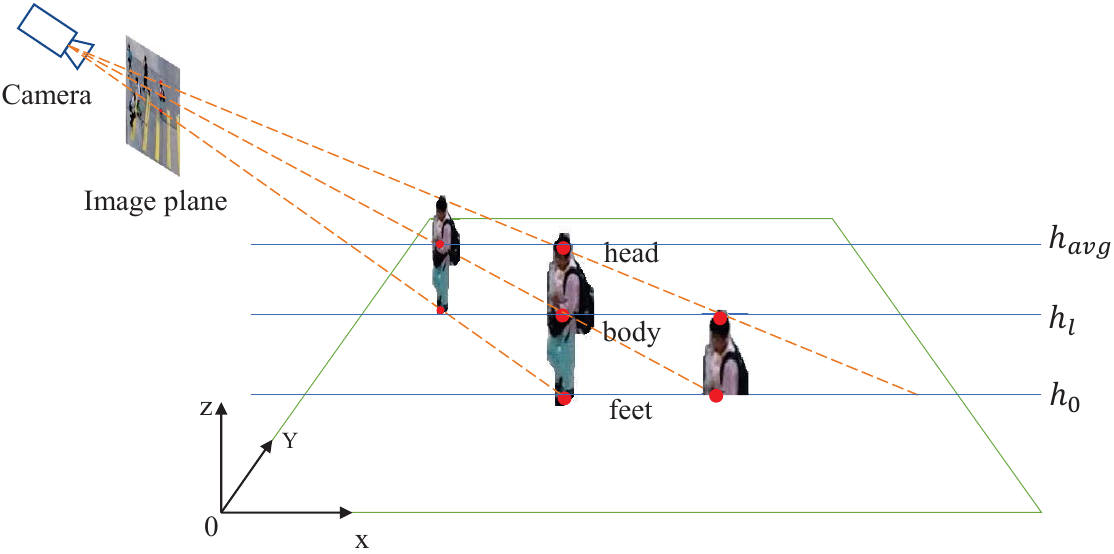}
   \caption{
   The multi-height projection can extract feature of a person along the $z$ dimension and form a 3D feature representation for the person, which is consistent with the 3D scene. }
\label{fig:3dprojection}
\end{figure}

\subsection{Single-view branch}
\par
Each camera-view image is fed into a CNN to extract the camera-view features. The 2D camera-view features are decoded by another CNN to predict the camera-view density maps. The camera-view feature extraction and density map prediction layer settings can be found in Table \ref{table:layer_setting} left.

\par
The camera-view prediction branches contribute to the model performance in two aspects. First, the camera-view prediction branches together with the 2D density map supervision improve the training of the camera-view feature extraction. A similar strategy is used by \cite{zhang2019wide}, where the camera-view prediction branches are used in the first stage for pre-training. Second, the camera-view prediction branches force the 2D feature representation to be consistent with the 3D feature representation, and the difference between them is the geometric projection. 
This constraint is natural because the 2D and 3D observations are linked by the geometric projection, and the same link still exists between the 2D and 3D feature representations. The 2D-3D feature representation constraint was also used by \cite{girdhar2016learning} for 3D reconstruction,
where the 3D representation is forced to be able to predict the corresponding 2D observations. Furthermore, this constraint can improve the final 3D prediction performance (see ablation study for more details).

\par \zq{Suppose} there are $N$ camera-views, the density map prediction for the $i_{th}$ view is $V_i\in {R^{h \times w}} $ and the corresponding ground-truth is $V_i^t \in {R^{h \times w}}$, then the camera-view density map prediction loss $l_{2d}$ is the mean-square error (MSE):
\begin{align}
  l_{2d} = \frac{1}{wh}\sum_{i=1}^N \parallel V_i - V_i^t \parallel _2^2.
\end{align}

\subsection{2D-to-3D projection}

\par
The extracted camera-view features are projected to the 3D world with a fixed 2D-to-3D projection layer. We assume that the camera intrinsic and extrinsic parameters are both known, which are used to perform the projection. Since each image pixel's corresponding height in the 3D world is unknown, a height range $H$
is used in the projection, where each pixel is projected to the 3D world multiple times onto different height planes. Then, the projected features from all heights are concatenated along the $z$-dimension 
to form a 3D feature representation.
The 2D-to-3D projection layer can be implemented using the Sampler from the Spatial Transformer Networks \citep{Jaderberg2015Spatial} with the known camera calibration parameters (intrinsic and extrinsic).
Suppose that $\cal P$ is the 2D-to-3D projection, $F_i$ is the 2D feature presentation of view $i$, and the height range $H = \{h_0, h_1, ..., h_r\}$, then the projected 3D feature representation $(F_{3d})_i$ for view $i$ is
\begin{align}
  (F_{3d})_i = {{\cal P}_i} (F_i, H)
             = [{{\cal P}_i} (F_i, h_0), ..., {{\cal P}_i} (F_i, h_r)] ,
\end{align}
where $[\cdot]$ is concatenation 
along the $z$-dimension (height).

The previous work of \cite{zhang2019wide} used the fixed-height projection, where all pixels were projected to the average person height (1750mm). Compared to the average-height projection, the multi-height projection can output a body feature representation along the $z$-dim (e.g., head, body and feet features, see Fig. \ref{fig:3dprojection}), which allows the network to better identify the location of the person through the 3D alignment of the features.

\begin{table}[t]
\centering
\scriptsize
\begin{tabular}{|c|c|}
\hline
\multicolumn{2}{|c|}{Attention along the view-ray} \\ \hline
Layer         & Filter      \\ \hline
\multicolumn{2}{|c|}{Feature extraction} \\ \hline
conv 1             & $64\! \times\! 1\! \times\!  3\!  \times\!  3$     \\ 
conv 2             & $64\!  \times\!  64\!  \times\!  3\!  \times\!  3$    \\ 
pooling   & $2\!  \times\!  2\!  $         \\ 
conv 3             & $128\!  \times\!  64\!  \times\!  3\!  \times\!  3$   \\ 
conv 4             & $128\!  \times\!  128\!  \times\!  3\!  \times\!  3$ \\ 
pooling   & $2\!  \times\!  2\!  $          \\ \hline
conv 5             & $256\!  \times\!  128\!  \times\!  3\!  \times\!  3$ \\ 
conv 6             & $256\!  \times\!  256\!  \times\!  3\!  \times\!  3$ \\ 
conv 7             & $256\!  \times\!  256\!  \times\!  3\!  \times\!  3$  \\ \hline

\multicolumn{2}{|c|}{Height map estimation} \\ \hline

conv 8             & $128\!  \times\!  256\!  \times\!  5\!  \times\!  5$ \\ 
conv 9             & $64\!  \times\!  128\!  \times\!  5\!  \times\!  5$ \\ 
conv 10             & $N\!  \times\!  64\!  \times\!  5\!  \times\!  5$  \\ 
softmax layer & - \\ 
resize layer & if applicable \\ \hline
\end{tabular}

\caption {The layer settings for the geometric attention module, including the feature extraction part and the height map estimation part. $N$ stands for the height level number in the height maps, which is 7 for PETS2009 and CityStreet, or 5 for DukeMTMC. For CityStreet, an extra resize layer is used to resize the height maps' $z$-dim to be 28; For DukeMTMC, zero padding along the $z$-dim is first adopted to increase height maps' $z$-dim size to be 9 and then resized to 36.
}
\label{table:attention_layer_setting}
\end{table}

\begin{figure*}[t]
\centering
   \includegraphics[width=0.9\linewidth]{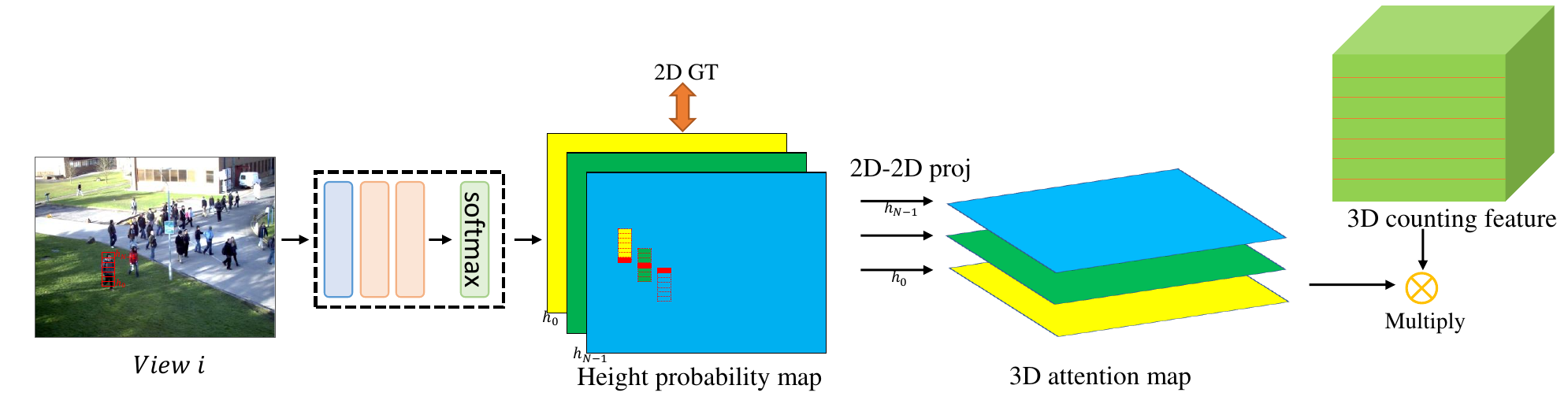}
   \caption{
   The geometric attention module. Each pixel in the camera view image is classified into $N$ height maps.  The $N$ height probability maps are projected to 3D scene \zqq{planes} according to their height levels \zqq{(a set of 2D-2D projections for different height levels)}. The projected height maps are used as attention to reduce the repeated the features in the 3D counting feature via multiplication.}
\label{fig:attention_module}
\end{figure*}

\zqb{\subsection{Geometric attention along the view ray}}

To enforce the geometric constraint that a 2D pixel should be projected to only one 3D voxel, we propose an attention module along the view ray. The attention module reduces the feature redundancy caused by copying the same 2D feature along the view ray in the typical 2D-to-3D projection step. The main idea is to project different body parts of the people to different height levels, which is different from projecting all to the same height level as in \citep{zhang2019wide}, and  projecting to all height levels like in 3D reconstruction.

A height prior can be incorporated in the model that the bottom body part is lower than the top part in the 3D world. Therefore, by dividing the human bounding boxes into several regions vertically, each region is assigned to a different height level. Note that the ground-truth bounding boxes are estimated from the people position annotations and camera calibrations \citep{zhang2019wide}, which does not need laborious annotation.  As shown in Figure \ref{fig:attention_module}, we divide each bounding box into $N$ parts, corresponding to $N$ height levels. Thus, a $ a \times b \times N$ binary map,
denoted as height maps ${\cal H}$, is constructed as supervisory information, where the element ${\cal H}_{ijh}$ indicates that the corresponding pixel $(i,j)$ \zqq{inside a bounding box} belongs to the corresponding height level $h$.
\zqq{For background pixels (not within any bounding box), the corresponding height map entires are set to 0.}

The geometric attention works as follows. First each 2D image pixel is classified into $N$ height levels by a height map estimation CNNs (\zqb{see the layer setting in Table \ref{table:attention_layer_setting}}). \zqq{Then, the 2D map of each height class is projected to a 3D scene plane at its corresponding height level.}
In this way, the geometric constraint is enforced along the view ray. Finally, the 3D height map serves as the attention by multiplying with the 3D crowd counting feature map.

Suppose the height map prediction is ${\cal H} \in {R^{a \times b \times N}} $ and the corresponding ground-truth is ${\cal H}^t \in {R^{a \times b \times N}}$, the height map prediction loss $l_{hmap}$ is the MSE
\begin{align}
  l_{hmap} = \frac{1}{abN} \parallel {\cal H} -  {\cal H}^t \parallel _2^2.
\end{align}


\subsection{3D density map prediction}

After the 2D-to-3D projection, the projected 3D features from multiple views are concatenated (along the feature channel) and then decoded by several 3D convolution layers.
The architecture for the 3D feature decoder layers can be found in the right of Table \ref{table:layer_setting}.
Let the 3D prediction be $G \in {R^{a \times b \times N}} $ and the corresponding ground-truth be $G^t \in {R^{a \times b \times N}}$. Then, the 3D prediction loss $l_{3d}$ is the MSE
\begin{align}
  l_{3d} = \frac{1}{abN} \parallel G - G^t \parallel _2^2.
\end{align}

\subsection{3D-to-2D projection consistency measure}
\par
Considering the geometric constraint between 2D camera plane and the 3D world, we also require the 3D prediction be consistent with the 2D single view density maps in terms of projection geometry. To achieve this, the 3D prediction $G$ is projected to each camera-view $i$ with a 3D-to-2D projection operation ${\cal P}^{'}$. The projection consistency between the projected 2D density maps and the 2D density map ground-truth is measured and used as part of the loss to further enhance the 3D counting performance.

\begin{figure*}[t]
\centering
   \includegraphics[width=1.5\columnwidth]{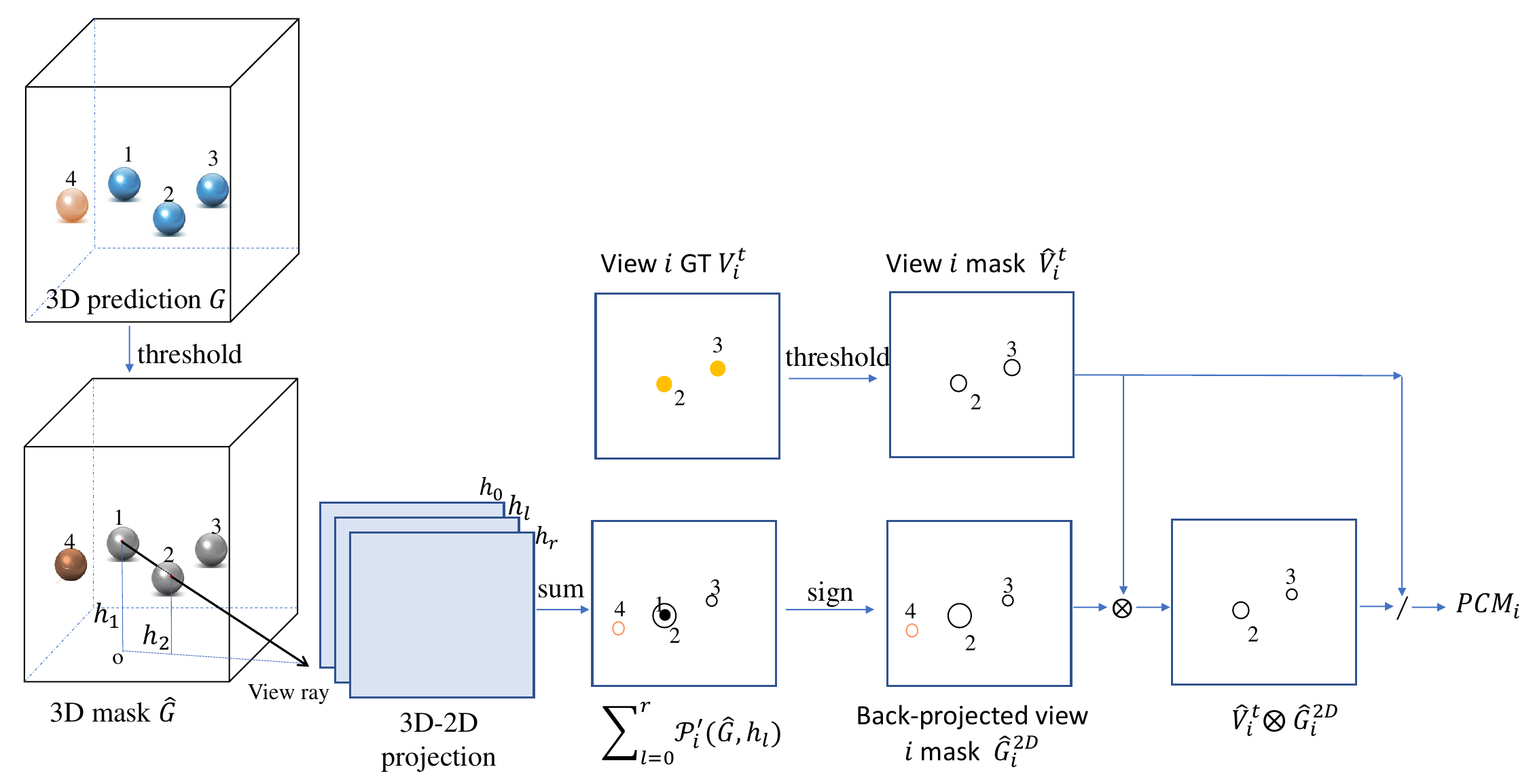}
   \caption{Example of projection consistency measure (PCM). \zq{There are 4 people in the 3D prediction, while only Person 2 and Person 3 are visible in the 2D view $i$. Since Person 1 is occluded by Person 2 and Person 4 is totally occluded in view $i$, they are masked out in the PCM calculation.} When the people location in the 3D prediction is not consistent with the 2D ground-truth, the $PCM$ value is low.
   }
\label{fig:pcm}
\end{figure*}

\textbf{3D-to-2D projection.}~Let $\cal P^{'}$ be the 3D-to-2D projection, and $\hat{G}$ be a 3D binary mask, which is the 3D prediction $G$ thresholded at $T=10^{-4}$. When performing the projection, without the height information, each 2D image point corresponds to many 3D points along the view ray. Therefore, the height range $H = \{h_0, h_1, ..., h_r\}$ is also used in the 3D-to-2D projection. Then the projected 2D mask $\hat{G}^{\mathrm{2D}}_{i}$ for view $i$ can be denoted as,
\begin{align}
      \zq{\hat{G}^{\mathrm{2D}}_{i}  = \sign(\sum^r_{l=0}{\cal P}^{'}_i (\hat{G}, h_l)),}
  \label{eqn:G2D}
\end{align}
where $\sum$ is summation along the $z$-dimension (height dimension). In the 3D-to-2D projection, the projected 2D pixel value can be regarded as the binary summation along the view ray in the 3D prediction grid (see Fig. \ref{fig:pcm}).

\textbf{Projection consistency measure.}~The main challenge to measure the consistency between the 3D density map prediction and the 2D camera-view density map is that some people who are occluded in a 2D camera-view will be present in the 3D prediction because they are visible from other views. Thus the MSE loss cannot be used directly here.
Instead, we measure the consistency between the binary masks produced by thresholding 
the 3D prediction and the corresponding  2D ground-truth density map, while accounting for inconsistencies due to occlusions (see Fig.~\ref{fig:pcm}).

\par
We define the 3D-to-2D projection consistency measure (PCM) for view $i$ as
\begin{align}
  PCM_{i} = \frac{\parallel \hat{V}_i^t \otimes \hat{G}^{\mathrm{2D}}_{i} \parallel_1}{\parallel \hat{V}_i^t \parallel_1 + \alpha},
\end{align}
where $\otimes$ denotes element-wise multiplication, $\hat{V}_i^t$ is a binary mask computed by thresholding the ground-truth 2D density map of view $i$ at $10^{-3}$,  $\hat{G}^{\mathrm{2D}}_{i}$ is the back-projected 3D prediction mask (see Eq.~\ref{eqn:G2D}), and $\alpha=10^{-5}$ is a constant to prevent divide-by-zero.
%
The important property of the PCM is that no penalty occurs when an extra person is present in the 3D prediction $\hat{G}^{\mathrm{2D}}_{i}$, but not the 2D camera-view (e.g., due to occlusion). On the other hand, the PCM will be reduced when a person in the 2D camera-view is missing in the 3D prediction. Finally, the projection consistency loss is defined by summing over the camera-views:
\begin{align}
  l_{3d\_2d} = \sum_{i=1}^{N}(1 - PCM_{i}).
\end{align}

\subsection{Training loss}

Combining all the aforementioned losses, the final loss is
\begin{align}
  l_{all} = l_{3d} + \beta l_{2d} + \gamma l_{3d\_2d}+ \sigma l_{hmap}, \label{loss_all}
\end{align}
where $\beta$, $\gamma$ and $\sigma$ are hyperparameters for weighting the contributions of each term.

\section{Experiments and Evaluation}
In this section, we will discuss the implementation details of the proposed 3D counting model, and conduct experiments on four multi-view counting datasets.

\begin{table*}
\small
\centering
\begin{tabular}{l|c|c|c}
\hline
    Dataset        & PETS2009      &DukeMTMC    &CityStreet   \\
\hline
    Dmap weighted  & 8.32                                      & 2.12                             & 9.36   \\
    Detection+ReID & 9.41                                       & 2.20                             & 27.60   \\
\hline
    Late fusion \citep{zhang2019wide}         & 3.92             & 1.27                             & 8.12    \\
    Na\"ive early fusion \citep{zhang2019wide} & 5.43             & 1.25                             & 8.10   \\
    MVMS \citep{zhang2019wide}                  &3.49              &1.03                     & 8.01   \\
\hline
    3D counting (ours)                        &\textbf{3.15}        & 1.01                          & 7.54 \\
    3D counting+Attention (ours)                        &3.20       &\textbf{0.92}                          & \textbf{7.12} \\
\hline
\end{tabular}
\caption{Experiment results: mean absolute error (MAE) on three multi-view counting datasets. ``3D counting'' uses both single-view branches and the projection consistency measure loss. ``3D counting+Attention'' adds the geometric attention module to the 3D counting model.}
\label{table:results}
\end{table*}

\subsection{Implementation details}

\textbf{3D ground-truth generation.}~The 3D ground-truth is generated with the single-view annotations (here we use head annotations) and the cross-view correspondence, \zqq{which are already available in the multi-view counting datasets}. Suppose the same person's corresponding annotations in $M$ views ($M$$\leq$$N$) are $\{(x_j, y_j)\}$, $j\in\{1, 2, ..., M\}$, their corresponding 3D coordinates are denoted as $\{W_j\} = \{{{\cal P}_j}((x_j, y_j, h))\}$ where the height $h$ is not known. Considering the projection correspondence, the 3D coordinate of the person's head $W=[X, Y, Z]^T$ can be obtained by finding the height $h$ that minimizes the spread of the projected head positions from all views,
\begin{align}
     &Z = \mathop{\mathrm{argmin}}_h \parallel W_j - \overline{W} \parallel^2_2, \label{Z} \\
     &(X, Y) = \frac{1}{M}\sum^M_{j=1}{{\cal P}_j}((x_j, y_j, Z)), \label{X_Y}
\end{align}
where $\overline{W}$ is the mean of $\{W_j\}$. $Z$ is the height $h$ minimizing (\ref{Z}), which can be obtained through searching $h$ in the height range $H' \in \{h'_0, h'_1, ..., h'_{r}\}$.
In our experiments we use range $\{1000, 1010, 1020, ... , 2000 mm\}$. 
For numerical stability during training, 2D density  ground-truth maps are scaled by $10^3$,  while the 3D ground-truth is scaled by $10^4$.
The predictions are scaled downward correspondingly for evaluation.

\textbf{Training settings and methods.}~In the proposed method (see Fig.~\ref{fig:pipeline}), besides the 3D counting prediction, there are two important modules: single-view density map prediction and the projection consistency measure module. In the ablation study, we show the influence of these two modules on the final 3D counting performance.
A 3-stage training is used for the proposed method.
\zq{In the first stage, $\beta=1$ means that the single-view 2D supervision is dominant to benefit the feature extraction training. In the second stage, $\beta=0.01$ means the 3D supervision is dominant  to accelerate the 3D fusion training. In the third stage,  $\beta=0.01$ and
$\gamma$ is variable according to ground-truth setting,
which increases the influence of the projection consistency measure (PCM)  
to further enhance the performance.} In all stages, the learning rate is set as $10^{-4}$ and the batch size is 1.

\subsection{Experiment setup}

\subsubsection{Datasets.}~The proposed method is evaluated and compared on the 4 multi-view counting datasets, PETS2009 \citep{ferryman2009pets2009}, DukeMTMC \citep{ristani2016MTMC}, CityStreet \citep{zhang2019wide}, and CVCS \citep{zhang2021cvcs}. We directly use the same dataset settings by \cite{zhang2019wide}.

\textbf{PETS2009} \citep{ferryman2009pets2009} contains 3 views,
and 1105 and 794 images are for training and testing, respectively.
\zq{The input image resolution ($w \! \times \! h \! \times \! d$) is  $384\! \times \!288$ and the 3D ground-truth resolution is $152\!\times\!177 \!\times\! 7$. The voxel height in $z$-dim is 40cm in the real world and the height range is 0 to 2.8m.}

\textbf{DukeMTMC} \citep{ristani2016MTMC} contains 4 views, and the first 700 images are used for training and remaining 289 images are for testing. \zq{The input image resolution is  $640\! \times \!360$ and the 3D ground-truth resolution is $160\!\times\!120 \!\times\! 36$. The voxel height in $z$-dim is 10cm in the real world and the height range is 0 to 3.6m. \abc{The larger height range is due to change of the ground height in the scene.}
}

\textbf{CityStreet} \citep{zhang2019wide} consists of 3 views and 500 images in which the first 300 are for training and remaining 200 for testing. The input image resolution is  $676\! \times \!380$ and the \zq{3D ground-truth resolution is $160\!\times\!192 \!\times\! 28$. The voxel height in $z$-dim is 10cm in the real world and the height range is 0 to 2.8m.}

\textbf{CVCS} \citep{zhang2021cvcs} is a large synthetic multi-view crowd counting dataset containing 31 scenes with around 100 camera views in each scene, where 23 scenes for training and 8 scenes for testing. The input image resolution is $640\! \times \!360$. The 3D ground-truth resolution is $640\! \times \!720 \! \times\! 7$, and patch-based training is used and the cropped ground-truth size is $ 80\! \times \! 90 \times 7$. The voxel height in $z$-dim is 40cm in the real world and the height range is 0 to 2.8m.

\begin{table}
\centering
\small
\begin{tabular}{l@{\hspace{0.1cm}}l|ccc@{\hspace{0.1cm}}c}
\hline
  &  $n*h$  & 3D     &3D+2D    & \multicolumn{2}{c}{3D+2D+PCM} \\
\hline
 &  7*40cm    & 4.12   &3.20   &\textbf{3.15} & ($\gamma=100$)     \\
PETS2009 & 14*20cm    & 4.88     & 4.57      & 4.24 &($\gamma=10$)  \\
&  28*10cm    & 5.34     & 4.27     & 4.21 &($\gamma=1$) \\
\hline
 &  9*40cm    & 1.82     &1.71     &1.65 &($\gamma=10$)      \\
DukeMTMC  &18*20cm    & 2.12     & 1.63    & 1.49 &($\gamma=3$)   \\
  &36*10cm    & 2.15     & 1.03    & \textbf{1.01} &($\gamma=0.1$)     \\
\hline
  & 7*40cm
             & 8.98     & 8.49      & 8.35 &($\gamma=100$)       \\
CityStreet  &14*20cm
             & 8.72    & 7.89      & 7.71 &($\gamma=30$)   \\
  &28*10cm
             & 7.87  & 7.58      & \textbf{7.54} &($\gamma=10$)\\
        \hline
\end{tabular}
\caption{Ablation study of the training loss and ground-truth settings. $\gamma$ is the hyperparameter for the projection consistency measure (PCM) loss. $n$ is the number of voxels in the $z$-dimension (height), and $h$ is the voxel height in the 3D world. The evaluation metric is MAE.}
\label{table:ablation_study_combo}
\end{table}

\subsubsection{Comparison methods.}~Five multi-view counting methods are used for comparison.
1) ``Dmap weighted'' fuses single-view density maps into a scene-level count with a view-specific weight map, which is constructed based on how many views can see a particular pixel. In the experiment, CSR-net \citep{li2018csrnet} is used to predict the single-view density maps for datasets PETS2009 and CityStreet, and FCN-7 for DukeMTMC.
2) ``Detection + ReID''  first detects all humans in each camera-view and then the scene geometry constraints and person ReID are used to associate the same people across views. Specifically, Faster-RCNN \citep{ren2015faster} is used for people detection and LOMO 2015 \citep{liao2015person} for person ReID.
3) ``late fusion'' model fuses single-view density maps to predict 2D height-plane density maps \citep{zhang2019wide};
4) ``na\"ive early fusion'' model fuses feature maps to predict 2D height-plane density maps \citep{zhang2019wide};
5) ``multi-view multi-scale (MVMS)'' model fuses feature maps with a scale selection module to cope with the scale variation issue \citep{zhang2019wide}.

\textbf{Evaluation metric.}~The mean absolute counting error (MAE) and normalized mean absolution error (NAE) are used to evaluate the scene-level counting performance, comparing the scene-level predicted and ground-truth counts:
\begin{align}
  \mathrm{MAE}&=\tfrac{1}{N}\sum\nolimits_{i=1}^N|\hat{c}_i-c_i|,\\
  \mathrm{NAE}&=\tfrac{1}{N}\sum\nolimits_{i=1}^N\frac{|\hat{c}_i-c_i|}{c_i},
\end{align}
where $c_i$ is the ground truth count and $\hat{c}_i$ is the predicted count.


\begin{figure*}[t]
\centering
   \includegraphics[width=0.9\textwidth]{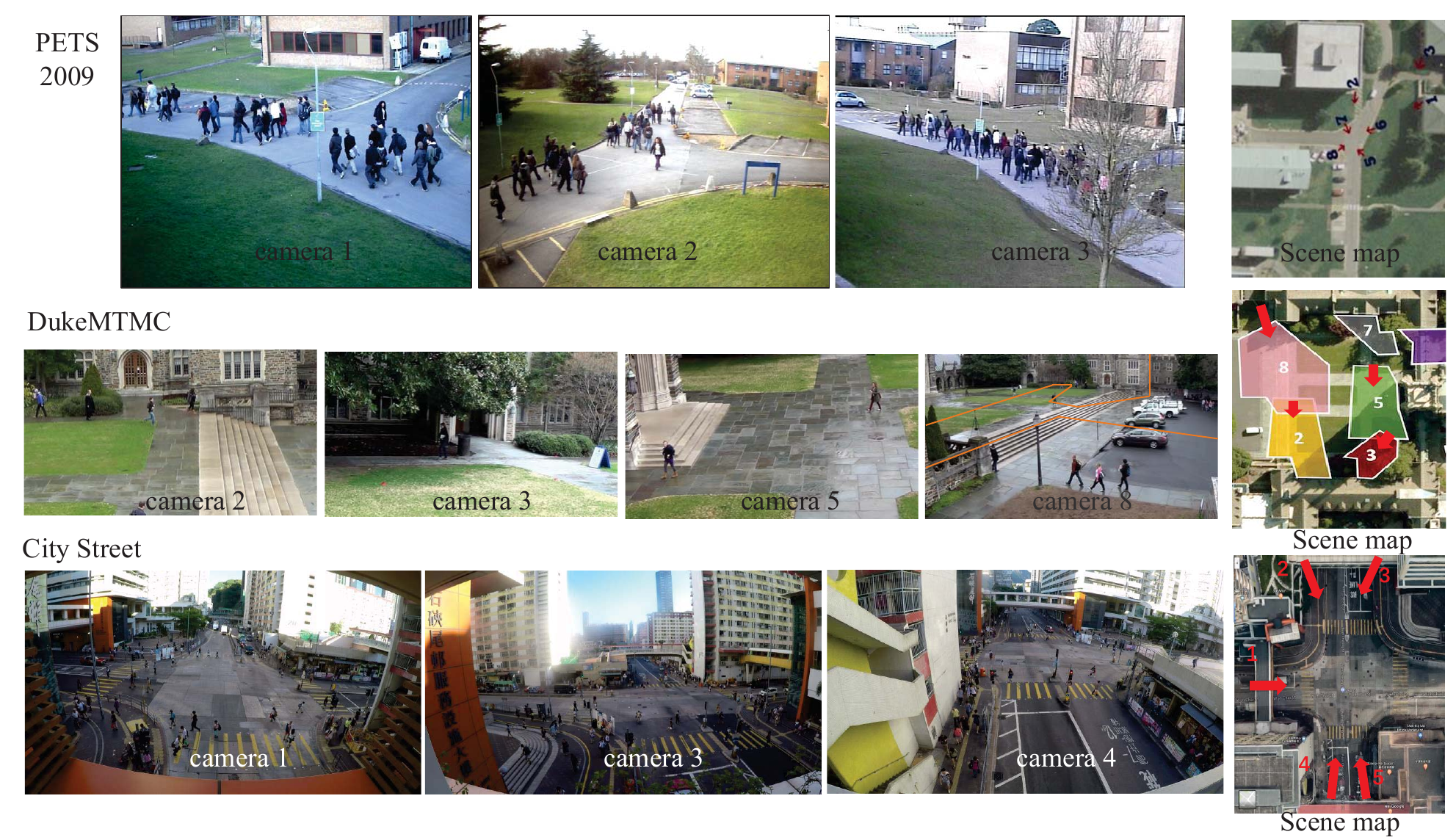}
   \caption{Examples of  3 multi-view datasets.}
\label{fig:datasets}
\end{figure*}

\subsection{Experiment results}

\begin{figure*}[t]
\centering
   \includegraphics[width=0.8\textwidth]{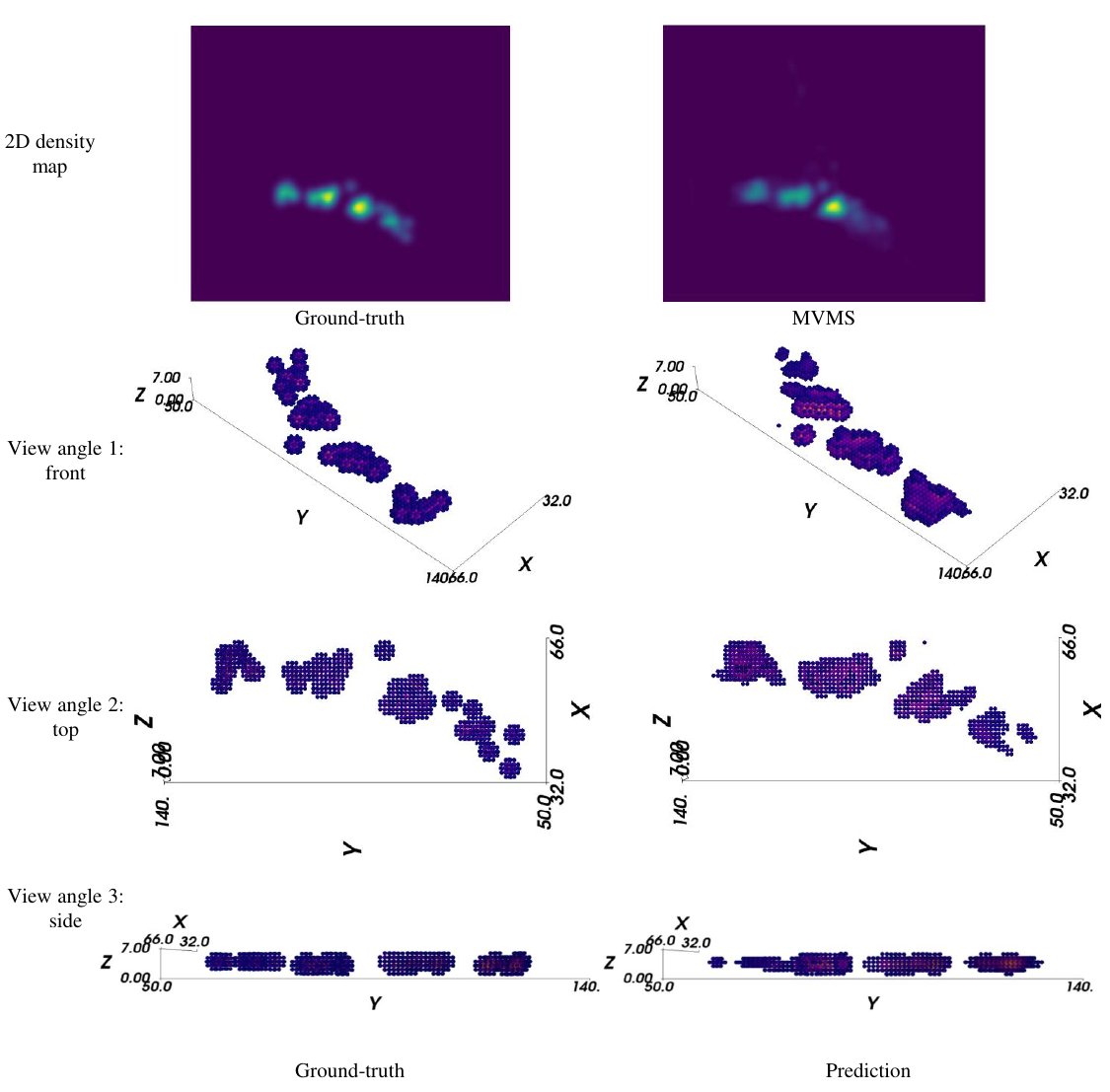}
   \caption{Examples of the ground-truth and the predictions on PETS2009.
   The 3D density maps of PETS2009 are thresholded by 5e-3, which are shown in 3 view angles: front, top and side. The 2D density map ground-truth and the prediction of the comparison method MVMS are also shown.
   }
\label{fig:PETS2009}
\end{figure*}

\begin{figure*}[t]
\centering
   \includegraphics[width=0.9\textwidth]{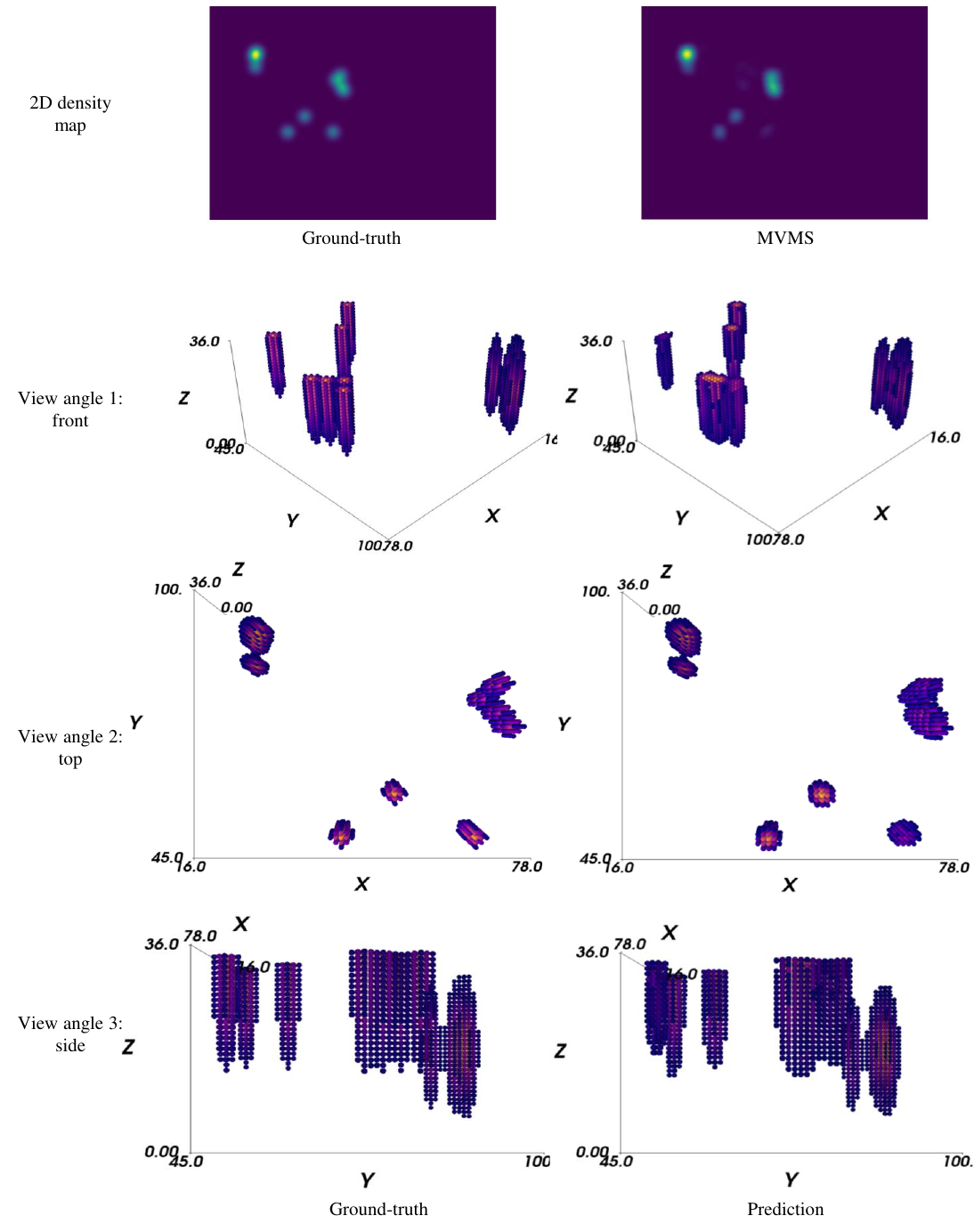}
   \caption{Examples of the ground-truth and the predictions on DukeMTMC.
   The 3D density maps of DukeMTMC are thresholded by 1e-3, which are shown in 3 view angles: front, top and side. The 2D density map ground-truth and the prediction of the comparison method MVMS are also provided. The third view angle (side) shows the variations in z-dimension of the people in the DukeMTMC dataset due to a staircase and raised ground-level.
   }
\label{fig:DukeMTMC}
\end{figure*}

\begin{figure*}[t]
\centering
   \includegraphics[width=0.9\textwidth]{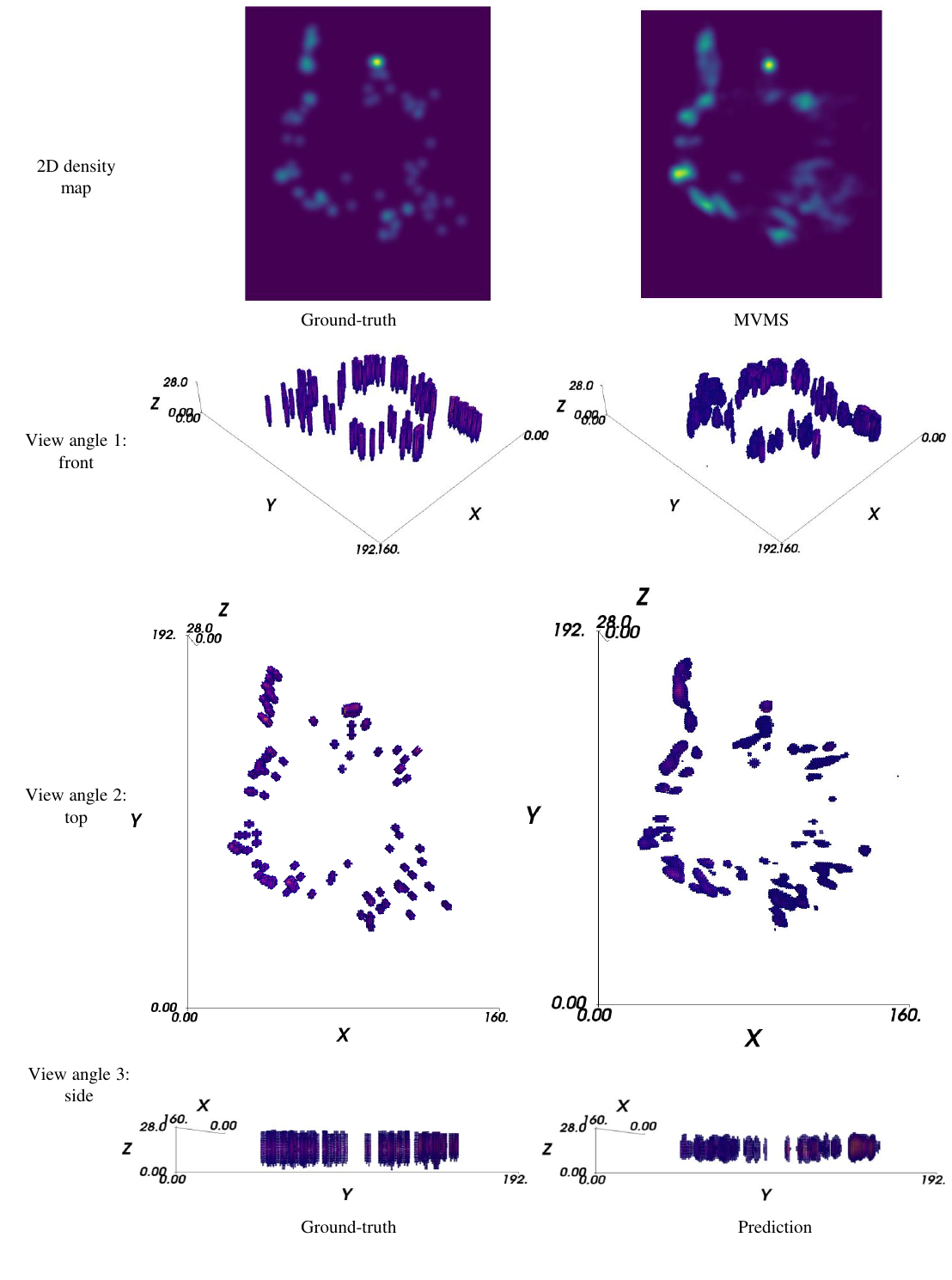}
   \caption{Examples of the ground-truth and the predictions on CityStreet.
   The 3D density maps of CityStreet are thresholded by 2e-3, which are shown in 3 view angles: front, top and side. The 2D density map ground-truth and the prediction of the comparison method MVMS are also shown.
   }
\label{fig:CityStreet}
\end{figure*}

The experimental results are shown in Table \ref{table:results} and the visualization results are found in the Figures \ref{fig:PETS2009}, \ref{fig:DukeMTMC} and \ref{fig:CityStreet}.

On \textbf{PETS2009}, the proposed 3D multi-view counting method can achieve better results than the two baseline multi-view counting methods (``Dmap weighted'', ``Detection + ReID'') and the 3 versions of the end-to-end multi-view counting method proposed by \cite{zhang2019wide}. The first two baseline methods cannot effectively fuse the multi-view information, which limits their performance. The proposed method achieves better performance than MVMS \citep{zhang2019wide}, which shows the advantage of the 3D projection and 3D fusion.

On \textbf{DukeMTMC}, our 3D multi-view counting method achieves slightly better performance than MVMS \citep{zhang2019wide}. The proposed method still achieves better performance than the two baseline methods. Due to low crowd count and lack of occlusions in the DukeMTMC, the performance gap is not very obvious.
\abc{Looking at the 3D predictions in Fig.~\ref{fig:DukeMTMC} (side view), our model can successfully predict people at different ground heights, which is consistent with the staircase and change in ground-level of the scene.}

On \textbf{CityStreet}, our 3D multi-view counting method achieves the best results among the end-to-end multi-view counting methods (late fusion, naive early fusion, MVMS) and the two baseline methods. ``Detection + ReID'' performs poorly on CityStreet due to large crowd count and severe occlusions.

On \textbf{CVCS}, the backbone model with our 3D fusion achieves better performance than the original backbone of \cite{zhang2021cvcs}. Since the 3D fusion is not proposed for cross-view cross-scene task, the performance is inferior to the CVCS model that contains extra modules for specifically handling cross-scene and cross-view data.
\abc{Integrating the CVCS modules with the 3D fusion is interesting future work.}

\begin{table}
\centering
%

\begin{tabular}{l|c|c}
\hline
    Method & {MAE}   & {NAE} \\
\hline
   Backbone                & 14.13      & 0.115    \\
   Backbone+3D fusion      & 13.30      & 0.123    \\
   \hline
   CVCS                    & 7.22      & 0.062    \\
\hline
\end{tabular}

\caption{The result comparison on the CVCS dataset. ``Backbone'' and ``CVCS'' are the models from \cite{zhang2021cvcs}, where ``Backbone'' is the baseline model without any other extra modules to handle the cross-scene or cross-view data. ``Backbone+3D fusion'' is the method replacing 2D fusion  in the ``Backbone'' model of \cite{zhang2021cvcs} with our 3D fusion.}
\label{table:result_cvcs}
\end{table}

\subsection{Ablation study}

In this section, we perform various ablation studies on the training loss,  the ground-truth settings, height map estimation, and geometric attention.

%

\begin{table*}
\centering
\small
\begin{tabular}{l|c@{\hspace{0.1cm}}c|c@{\hspace{0.1cm}}c|c@{\hspace{0.1cm}}c}
\hline
    Dataset & \multicolumn{2}{c|}{PETS2009}   & \multicolumn{2}{c|}{DukeMTMC}    & \multicolumn{2}{c}{CityStreet}\\
\hline
    Geometric Attention   &3D+2D    & \multicolumn{1}{c|}{3D+2D+PCM}       &3D+2D    & \multicolumn{1}{c|}{3D+2D+PCM}     &3D+2D     & \multicolumn{1}{c}{3D+2D+PCM}  \\
\hline
   without     &3.20   &\textbf{3.15}  ($\gamma=100$)
                 & 1.03     & 1.01 ($\gamma=0.1$)
                 & 7.58      & 7.54 ($\gamma=10$)\\
  with        & \textbf{3.15}      & 3.20 ($\gamma=100$)
                 & 0.94    & \textbf{0.92} ($\gamma=0.1$)
                & 7.15     & \textbf{7.12} ($\gamma=10$)   \\
\hline
\end{tabular}
\caption{Ablation study: MAE for the proposed method with and without the geometric attention module. Two kinds of loss functions are used with/without the geometric attention: 3D+2D and 3D+2D+PCM losses. The parameter $\sigma$ of the height map estimation loss $l_{hmap}$ is 1000 for CityStreet, 5000 for PETS2009, and 10 for DukeMTMC.
}
\label{table:ablation_study_attention}
\end{table*}

\subsubsection{Training loss.}
We first compare training with different losses.
The columns of Table \ref{table:ablation_study_combo} show the results of using 3D loss, 3D+2D loss or 3D+2D+PCM loss on the 3 datasets.
Using single-view prediction branches and 2D supervision (3D+2D loss) can achieve better multi-view counting performance in comparison with only using 3D loss.
Furthermore, using 3D+2D together with PCM loss can obtain better multi-view counting performance compared to using only 3D loss or 3D+2D loss on all 3 datasets with different ground-truth settings.

\zqq{
We next perform an ablation study on the weights of the loss terms on the  PETS2009 and CityStreet datasets. The final loss contains 3D counting and 2D counting losses, height map estimation loss and projection consistency loss.
The 2D counting loss is used to encourage extraction of useful counting features during training of the 3D counter. For simplicity, we use 0.01 as the 2D counting loss weights for all datasets.
Since we use a pre-trained counter, then the 2D count loss is used to keep the features informative for 2D counting.}

\zqq{
For the height map estimation loss, Table \ref{table:sigma} shows the results using different values of weight $\sigma$. When $\sigma$ is too small, the height estimation module is not well trained, and the model performance is limited. In contrast, when $\sigma$ is too large, the 3D counting estimation is not well trained, so the performance drops. In the end, we select $\sigma = 5000$ 
 for PETS2009 and $\sigma = 1000$ for CityStreet.
For the projection consistence loss (PCM), we  perform an ablation study on the weight $\gamma$ in Table \ref{table:gamma}. When $\gamma$ is too small, the model performance is limited because PCM loss has little influence. When $\gamma$ is too large, the error caused by the projection (eg. calibration error) is introduced in the model training.  We select $\gamma = 100$ for PETS2009 and $\gamma = 10$ for CityStreet.
}

\begin{table}
\centering
\begin{tabular}{l|ccccc}
\hline
    $\sigma$                 & 	100	& 1000	& 5000	& 10000	& 20000 \\
\hline
   PETS2009               & 3.81	& 3.58	& 3.15	& 3.40	& 5.66   \\
   CityStreet             & 8.07	& 7.15	& 7.50	& 7.80	& 7.82    \\

\hline
\end{tabular}
\caption{Ablation study on the weight $\sigma$  for the height map estimation loss on PETS2009 and CityStreet dataset.}
\label{table:sigma}
\end{table}

\begin{table}
\centering
\begin{tabular}{l|ccccc}
\hline
    $\gamma$                 &	1	& 10	& 100 &	200 & 500 \\
\hline
   PETS2009               & 	3.45 &	3.52 &	3.20	&3.30	&3.59   \\
\hline
    $\gamma$                   &	1	&5	& 10	&50	&100 \\
\hline
   CityStreet               & 7.22	&7.27	&7.12	&7.21	&7.33   \\

\hline
\end{tabular}
\caption{Ablation study on weight $\gamma$  for the height map estimation loss on PETS2009 and CityStreet dataset.}
\label{table:gamma}
\end{table}

\subsubsection{Ground-truth setting.}
\label{sec:gtablation}
The columns of Table \ref{table:ablation_study_combo} show the results of using different resolution of the 3D density map ($n$ is the number of voxels in the $z$-dimension, and $h$ is the voxel height in the 3D world).
For PETS2009, the best performance is achieved by using 7 voxels and voxel height 40cm with 3D+2D+PCM loss ($\gamma=100$). For DukeMTMC, using voxel number 36 and voxel height 10cm with 3D+2D+PCM loss ($\gamma=0.1$) gives the best result. As to CityStreet, the best result is obtained by using voxel number 28 and voxel height 10cm with 3D+2D+PCM loss ($\gamma=10$). \zq{Compared to DukeMTMC and CityStreet, the best performance of PETS2009 is achieved at $h$=40cm. The people occlusion in PETS2009 is more severe and many people's lower bodies are totally occluded from all views (e.g., the people in the middle).
Thus, increasing the height resolution does not provide additional information of the body, but may introduce more noise (other people's features) along the $z$-dim, thus leading to worse performance.}

\zqb{\subsubsection{Geometric attention.}}
We compare the 3D counting model's performance with/without  the module for attention along view ray in Table \ref{table:ablation_study_attention}. In addition to the height map estimation loss $l_{hmap}$, two kinds of loss combinations are considered, 3D+2D and 3D+2D+PCM losses. The best ground-truth settings are used for each dataset as indicated by the ablation study in Section \ref{sec:gtablation}. 
 For the loss $l_{hmap}$, the weight $\sigma=1000$ for CityStreet, $\sigma=5000$ for PETS2009, and $\sigma=10$ for DukeMTMC.

Results in Table \ref{table:ablation_study_attention} show that the performance can be improved with the geometric attention module. Especially on the DukeMTMC dataset, the performance gain of the model with geometric attention module is significantly improved (\zqq{about $10\%$}) compared to results without the module. \zqb{The reason is the height change in DukeMTMC is larger than in other datasets and the geometric attention selects more suitable height levels for fusion, benefitting the multi-view fusion.}
\zqq{
On the CityStreet dataset, the performance with geometric attention is better than without using it, with a  gain of about $6\%$.
On the PETS2009 dataset, since the people height variance is smaller and most people are occluded in the scene, the effect of geometric attention is not as obvious, albeit it still achieves slightly better performance when without PCM loss.}


\zqq{\subsubsection{Height map estimation loss.}}

\zqq{
We have also performed an ablation study on the form of the height map estimation loss, either MSE or Cross Entropy (CE) loss, and whether to use a mask to only compute the loss in the people's bounding boxes (i.e., ignoring background pixels). The results in Table \ref{table:loss}
show that the model performance with CE loss is not better than with MSE loss. Furthermore, keeping background pixels in the predicted geometric attention maps can achieve better performance because the background can provide context information for multi-view fusion (to associate the same people or distinguish different people) in the next stage.}

\begin{table}
\centering
\begin{tabular}{l|cc}
\hline
    Method    &	PETS2009	& CityStreet \\
\hline
     No Mask + MSE                	&{\bf 3.15}	&{\bf 7.15} \\
   No Mask + CE               	&3.99	&9.25   \\
     Mask + MSE                 	&4.17	&7.72 \\
   Mask + CE               &3.56	&8.27   \\

\hline
\end{tabular}
\caption{Ablation study on the loss for the height map estimation module (no PCM loss). ``Mask'' means remove the background in the height map estimation loss and ``No Mask'' means keep the background in the loss. ``CE'' denotes cross entropy loss.}
\label{table:loss}
\end{table}

\zqq{\subsubsection{Geometric attention vs. spatial attention.}}
\zqq{
We have performed ablation experiments to show the effect of the height map supervision and compared with a common spatial attention module (w/ or w/o supervision) in Table \ref{table:attention}. The spatial attention module estimates the 2D occupancy maps of the crowd and then projects them to the 3D world as attention. From Table \ref{table:attention}, the performance of the proposed geometric attention module is better than the spatial attention module (regardless of with or without supervision).
Furthermore, without the height map supervision, the performance of the geometric attention module drops, because the height map estimation needs extra information to distinguish the body parts of the people which projected to different height levels.
}

\begin{table}
\centering
\begin{tabular}{l|cc}
\hline
    Method    &	PETS2009	& CityStreet \\
\hline
    Geometric Attention (w/)                	&{\bf 3.15}	&{\bf 7.15} \\
   Geometric Attention (w/o)               &4.59	&7.59  \\
    Spatial Attention  (w/)              &4.03	& 9.93 	 \\
    Spatial Attention (w/o)                &5.03	& 8.08 \\
\hline

\hline
\end{tabular}
\caption{Ablation study on the attention module (no PCM loss). ``w/'' and ``w/o'' indicate  with supervision or without attention supervision, respectively.}
\label{table:attention}
\end{table}

\section{Conclusion and Discussion}
\par In this paper, a DNN-based 3D multi-view counting method is proposed, which fuses camera-views to predict the 3D scene-level density map. 3D projection and fusion are used, which can handle the situation when people are not all located at the same height (e.g., 
people standing on a staircase),
and provides a chance to solve the scale variation issue in the 3D space without a scale selection operation. The projection consistency measure between the 3D prediction and 2D density map ground-truth is studied and then utilized in the loss function to refine the 3D prediction further.
Compared to other state-of-art multi-view counting methods, the proposed method  achieves better or comparable counting performance as well as a more informative scene-level crowd representation.

In addition to counting humans, the proposed 3D multi-view counting method can also be applied to counting birds in the sky or the fish in the aquarium, where both the bird or the fish count can be obtained as well as their 3D location distributions -- of course, this requires collecting more multi-view scenes. In addition to  object counting, since the 3D Gaussian kernels are used as ground-truth, the 3D prediction provides a vivid visualization for the scenes, as well as the potentials for other applications like observing the scene in arbitrary view angles, which may contribute to better scene understanding, generation or visualization.

%

\begin{acknowledgements}
This work was supported by grants from the Research Grants Council of the Hong Kong Special Administrative Region, China (CityU 11212518, CityU 11215820), and by a Strategic Research Grant from City University of Hong Kong (Project No. 7005665).
\end{acknowledgements}

\bibliographystyle{spbasic}      
\bibliography{egbib}   

\end{document}